# Enhancement of Noisy Planar Nuclear Medicine Images using Mean Field Annealing

D.L. Falk[1], D. M. Rubin[1] and T. Marwala[1]

[1] School of Electrical and Information Engineering, University of the Witwatersrand, Johannesburg, South Africa

*Abstract*— **Nuclear medicine (NM) images inherently suffer from large amounts of noise and blur. The purpose of this research is to reduce the noise and blur while maintaining image integrity for improved diagnosis. The proposed solution is to increase image quality after the standard pre- and post-processing undertaken by a gamma camera system. Mean Field Annealing (MFA) is the image processing technique used in this research. It is a computational iterative technique that makes use of the Point Spread Function (PSF) and the noise associated with the NM image. MFA is applied to NM images with the objective of reducing noise while not compromising edge integrity. Using a sharpening filter as a post-processing technique (after MFA) yields image enhancement of planar NM images.**

*Keywords*— **Mean Field Annealing, Nuclear Medicine, Image Restoration**

## I. INTRODUCTION

The MFA algorithm makes use of two techniques to achieve image restoration. Gradient descent is used as the minimization technique, while optimization is achieved by a deterministic approximation to Simulated Annealing (SA) [1]. The algorithm *anisotropically diffuses* an image, iteratively smoothing regions that are considered non-edges while still preserving edge integrity until a global minimum is obtained. A known advantage of MFA is that it is able to minimize to this global minimum skipping over local minima while providing comparable results to SA with significantly less computational effort [2].

Image blur is measured using either a point or line source. Both allow for the derivation of a PSF which is used in the image de-blurring process. The noise variance can be measured using a flood source. Noisy blurred NM images can be difficult to diagnose particularly at edges, for this reason MFA is suitable for planar NM image restoration.

Planar NM images are assumed to be piece-wise continuous images and thereby produce Markovian neighborhoods and allows the representation of images as Markov Random Fields [3].

## II. MFA APPLIED TO NUCLEAR MEDICINE IMAGES

### A. Theory

A Bayesian approach is used to define the objective function. Consider an ideal image $f$ and a measured image $g$; the a-posteriori conditional probability is given by:

$$P(f|g) = P(g|f)P(f)/P(g) \quad (1)$$

where

$$P(g|f) = \prod_{i,j} \frac{1}{\sqrt{2\pi}\sigma} \exp\left[-\frac{\left((f \otimes h)_{i,j} - g_{i,j}\right)^2}{2\sigma^2}\right] \quad (2)$$

and

$$P(f) = \prod_{i,j} \exp\left[\frac{1}{T} \exp\left(\frac{-\Lambda^2_{i,j}}{2T^2}\right)\right] \quad (3)$$

Where $T$ is referred to as the temperature and $\Lambda$ is a gradient operator. The quadratic variation operator (for more information see Wang et al. [1]) is used as the gradient operator in this particular algorithm but it is likely that more suitable Gradient Operators are available for various types of NM images. The aim is to seek an estimate of $f$ referred to as $f^*$ which will maximize the posterior conditional probability [4].

The objective function for minimization is derived by substituting Equations 2 & 3 into 1, taking the natural logarithm, changing the sign and ignoring the constant term. The objective function is the Hamiltonian and becomes a function of the image estimate $f^*$:

$$H(f^*) = \sum_{i,j} \frac{1}{2\sigma^2}\left((f^* \otimes h)_{i,j} - g_{i,j}\right)^2 + \beta \sum_{i,j} -\frac{1}{T}\exp\left(\frac{-\Lambda^2_{i,j}}{2T^2}\right) \quad (4)$$

Equation 4 ensures that the restored image has the maximum probability of being in the form of $f$ while still minimizing the noise in $f^*$ by penalizing and hence smoothing it. The coefficient β in front the prior Hamiltonian term is the ratio between penalizing noise and maintaining form. The prior term, also referred to as the penalty function, takes the form of an inverted Gaussian function. As the temperature $T$ decreases, penalty increases as can be in seen Fig. 1.

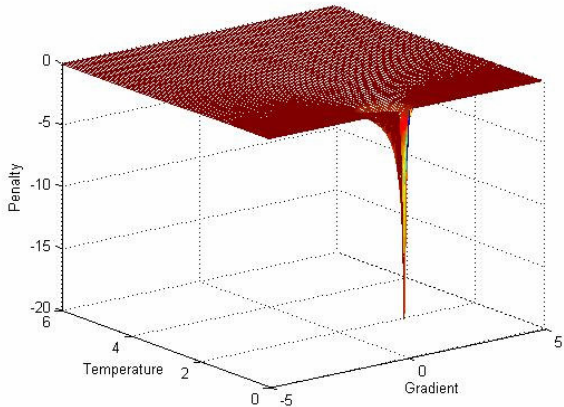

Fig. 1 Depiction of the inverted Gaussian as the penalty function.

After the objective function has been derived, gradient descent is utilized for the minimization process.

$$f_{i,j}^{k+1} = f_{i,j} - \alpha \frac{\partial H_T(f^k)}{\partial f_{i,j}} \quad (5)$$

Detailed discussion for the determination of value of α and the partial derivative required in gradient descent may be found in Bilbro et al. [5]. An outline of the MFA iterative technique can be seen in Fig. 2.

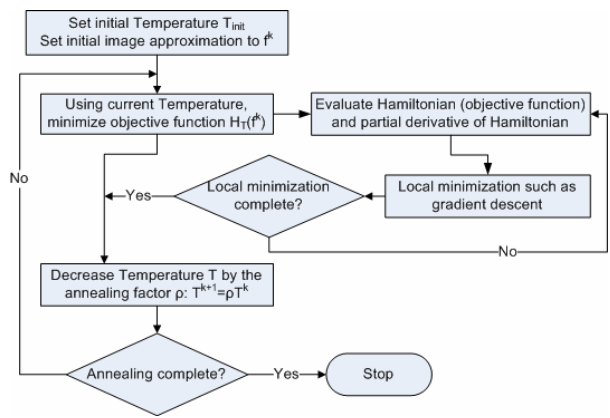

Fig. 2 Flow chart depicting overview of MFA algorithm.

*B. Discussion*

A vital aspect of MFA is defining when the annealing process is complete. Excessive annealing will add significant blur to the image. Normally when dealing with phantom images error metrics such as PSNR or RMSE may be used since the algorithm may constantly compare the restored image with the real image. However, when dealing with NM images this comparison is can not be made. The easiest solution would be to provide NM physicians with a "movie" of the restoration process and allow the NM physician to view the iterated image of choice. A more mathematical approach at achieving the correct stopping criteria is suggested by using the noise and prior Hamiltonians as enhancement indicators.

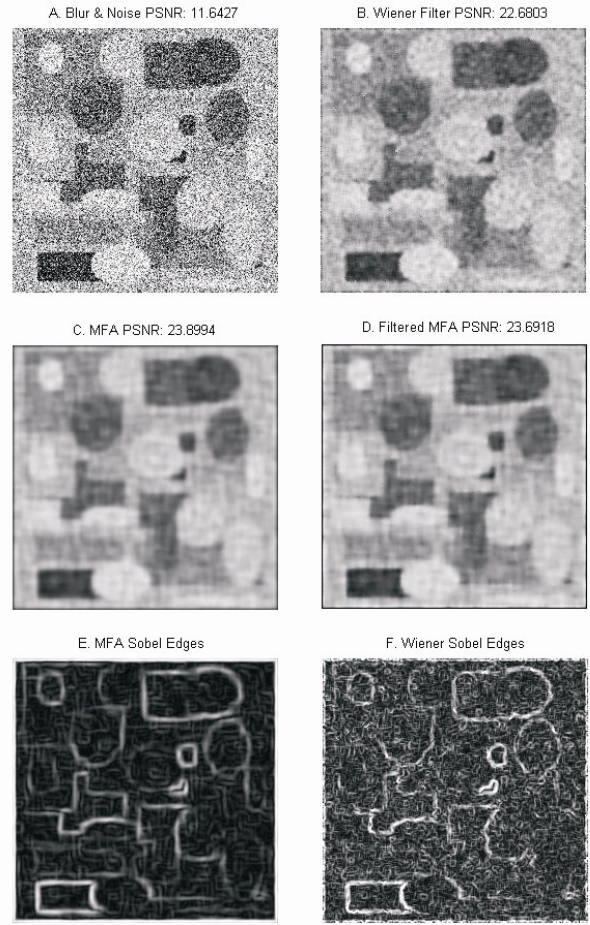

Fig. 3 Comparison between MFA and Wiener phantom results.

Phantom images were used extensively in the development of a MFA algorithm and MFA parameters. Experimental empirical methods were used on numerous phantom images such as Fig. 3 to determine optimal parameters.

It is evident from Fig. 3A & C that the MFA algorithm with the correct parameters can reduce noise substantially without damaging edge integrity. Fig. 3B shows a Wiener filter restored image. Comparative noise reduction and edge classification is evident from Fig. 3E & F, which displays the Sobel edges.

Looking carefully at Fig. 3A, B & C, it is noticeable that edges appear sharper in the original image and Wiener image in certain regions compared to the MFA restored image. This implies that MFA has blurred the image slightly. However since MFA has extensively reduced the noise it is now possible to apply filters to further enhance image edges without out amplifying the noise. A standard sharpening filter $h$ is used and the result is visible in Fig. 3D.

$$h = \begin{bmatrix} -0.167 & -0.67 & -0.167 \\ -0.67 & 4.33 & -0.67 \\ -0.167 & -0.67 & -0.167 \end{bmatrix} \quad (6)$$

This post-sharpening affect is more evident in the clinical results (see Section III) and highlights the strengths of MFA, which can be categorized as a supplementary pre-filter image enhancing technique.

One of the disadvantages in NM as shown Fig. 4 is that blurring increases with increasing distance from the collimator i.e. it is not depth invariant. This implies that the PSF will only be correctly specified for a single plane in the image. The other planes will experience a blurring affect due to the MFA process and an incorrectly specified PSF. NM physicians can therefore select *planes of interest* by modifying the Gaussian distribution to define the PSF used in MFA. The *planes of interest* will experience image enhancement while the other planes may experience increased blur.

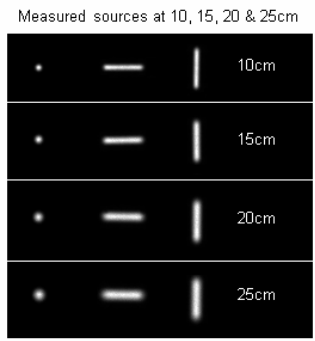

Fig. 4 Point and line sources at 10, 15, 20, 25cm.

To determine the PSF, point sources were placed at various distances away from the collimator. A discrete Gaussian distribution was then fitted to the acquired point source. Vertical and horizontal line sources were imaged using capillary tubes to verify the point sources' distributions and to verify the radial symmetry of the blur. Fig. 5 shows how the point source is convolved with a line and then compared to the acquired line source. Ignoring ends, the two lines suffered only small differences with an RMSE of 5.5% that may be attributed to the noise. The process is repeated with the vertical line resulting in a RMSE of 4.8% which implies approximate radial symmetry. Radial symmetry and the fitted Gaussian PSF were verified at numerous distances. Fig. 6 shows the Standard Deviation of the resulting fitted PSFs, in which the PSFs display regional linearity. A linear trend line may be fitted and used to predict approximate PSFs at different distances from the collimator.

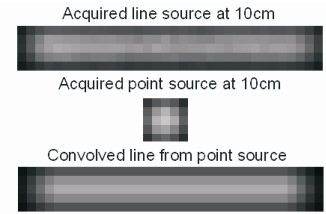

Fig. 5 Comparison of convolved line and acquired line source.

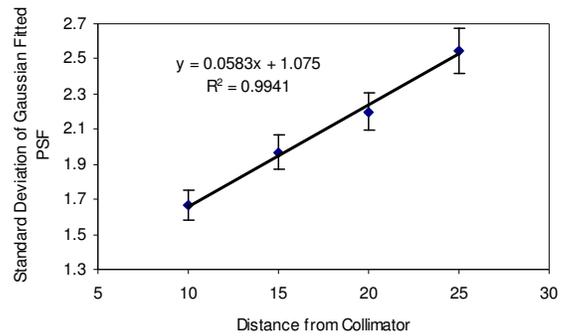

Fig. 6 Standard deviation of PSF measurements.

To simplify this work, PSF shift invariance is assumed, even though this is not strictly the case. A more complex algorithm may be derived to include a shift variant PSF.

III. RESULTS

It was determined experimentally that MFA parameters for NM images are similar to those used in Fig. 3C although they are not necessarily the optimal parameters). In this study the distance of the subject from the collimator is unknown, so a PSF of a standard deviation of 2 is chosen (corresponding to ±17cm from the collimator see Fig. 6). The noise variance is determined using a flood source and does not change with changing collimator distance. Note that noise variance does change with image intensity scaling and this change must be factored in when applying MFA.

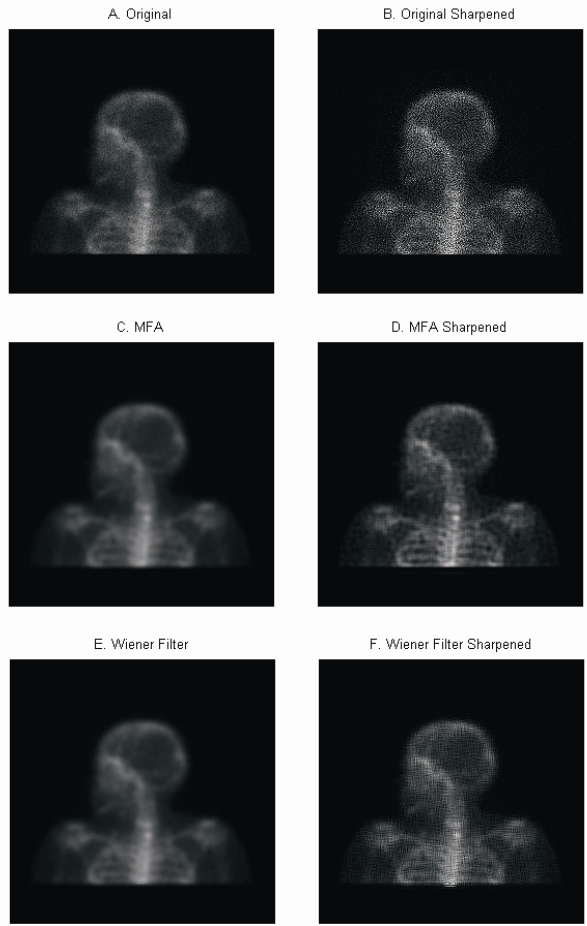

Fig. 7 Example of clinical results.

Fig. 7A shows an image acquired on General Electric Infinia gamma camera. Fig. 7C shows the restored image after 20 MFA iterations, the image appears to be slightly blurred with substantial reduction in noise. Fig. 7E shows a Wiener restored image which looks very similar to the MFA restored image. The images are all optimally sharpened using Equation 6 as the sharpening filter. The sharpening filter amplifies the noise in the original image as seen in Fig. 7B. The sharpening filter is run three times on the MFA restored image with image enhancement before noise amplification becomes apparent (Fig. 7D). The result is a clearer and sharper image which may improve diagnosis. In contrast the sharpening filter can only be run twice before noise amplification becomes visually obstructive in the Wiener restored image seen in Fig. 7F. Fig. 7D clearly contains more viewable detail then Fig. 7F.

## IV. Conclusions

With current processing technology the computational time required to run this image enhancing MFA algorithm is no longer significant. Although not all the criteria for image enhancement are present in NM images, enhancement of individual or multiple *planes of interest* is possible. Sharpening filters are utilized as a post-MFA enhancement technique and provide good results. We thus conlude that MFA holds promise as a supplementary pre-filter diagnostic tool for the enhancement of NM images.

## Acknowledgment

The authors would like to thank Prof. Vangu of the Department of Nuclear Medicine at Wits University for providing the research facilities required in this study. In particular, the authors would like to thank Mr. Sibusiso Jozela of the Medical Physics Department, for all his time spent acquiring the experimental data, and Mr. Nico van der Merwe, also of the Medical Physics department for all his input. We look forward to working with these departments to further develop this study.

Address of the corresponding author:

Author: Daniyel Falk
Institute: University of the Witwatersrand, Johannesburg
Street:
City: Johannesburg
Country: South Africa
Email: daniyelfalk@gmail.com